# SplatSearch: Instance Image Goal Navigation for Mobile Robots using 3D Gaussian Splatting and Diffusion Models

Siddarth Narasimhan, Matthew Lisondra, Haitong Wang, *IEEE Student Member*, Goldie Nejat, *IEEE Member*

*Abstract*— **The Instance Image Goal Navigation (IIN) problem requires mobile robots deployed in unknown environments to search for specific objects or people of interest using only a single reference goal image of the target. This problem can be especially challenging when: 1) the reference image is captured from an arbitrary viewpoint, and 2) the robot must operate with sparse-view scene reconstructions. In this paper, we address the IIN problem, by introducing SplatSearch, a novel architecture that leverages sparse-view 3D Gaussian Splatting (3DGS) reconstructions. SplatSearch renders multiple viewpoints around candidate objects using a sparse online 3DGS map and uses a multi-view diffusion model to complete missing regions of the rendered images, enabling robust feature matching against the goal image. A novel frontier exploration policy is introduced which uses visual context from the synthesized viewpoints with semantic context from the goal image to evaluate frontier locations, allowing the robot to prioritize frontiers that are semantically and visually relevant to the goal image. Extensive experiments in photorealistic home and real-world environments validate the higher performance of SplatSearch against current state-of-the-art methods in terms of Success Rate and Success Path Length. An ablation study confirms the design choices of SplatSearch. Our project page can be found at [https://anon-splatsearch.github.io/](https://anon-splatsearch.github.io/).**

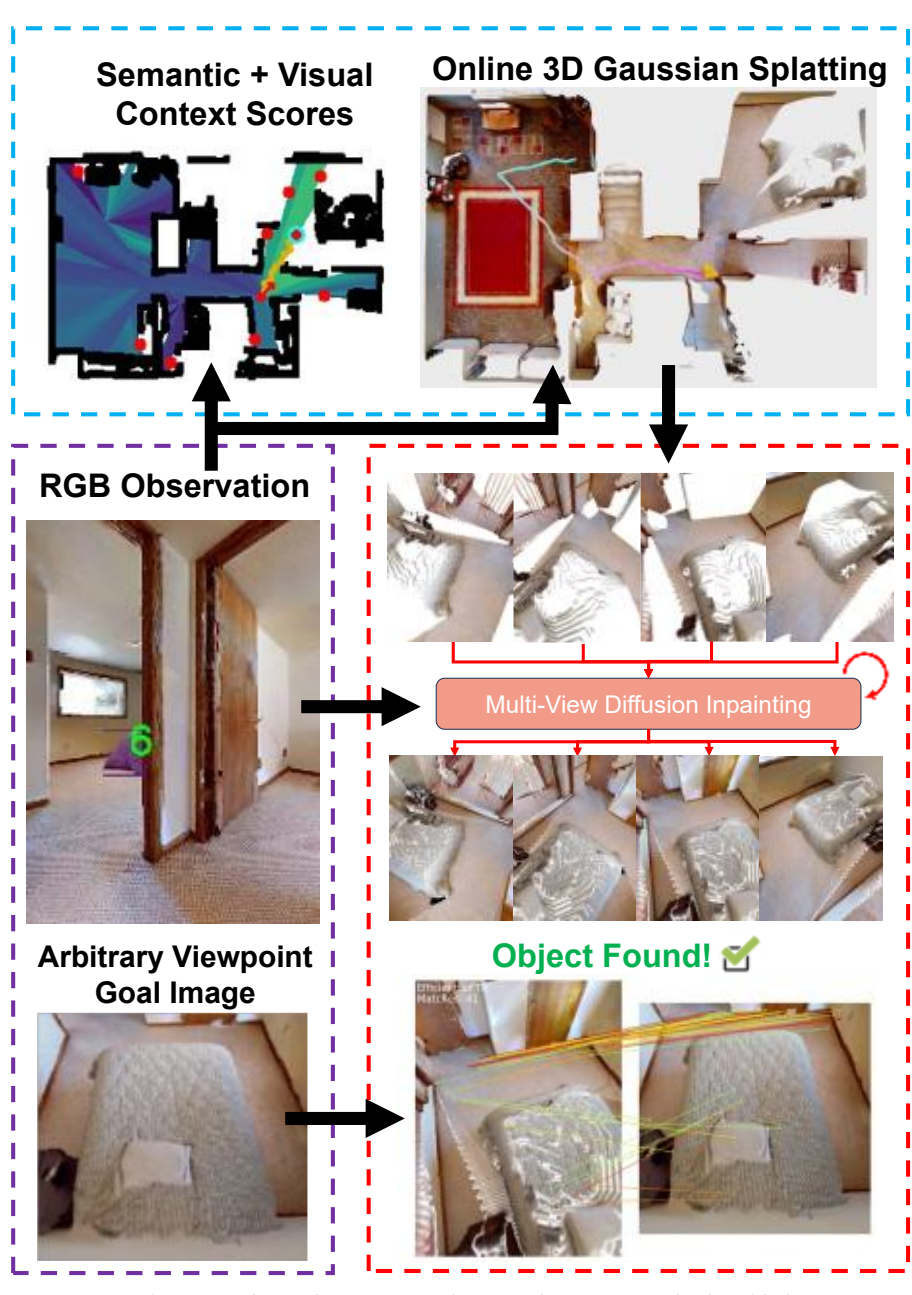

**Fig. 1.** An overview of SplatSearch. SplatSearch builds a sparse view 3DGS map and uses semantic and visual context scores to select frontier locations during exploration. To recognize goal images provided from arbitrary viewpoints, SplatSearch renders novel viewpoints around the objects and inpaints incomplete regions using a multi-view diffusion model.

## I. INTRODUCTION

Mobile robots are increasingly deployed in human-centered environments, where many tasks involve locating and navigating to people or objects of interest. In healthcare settings, robots may search for and deliver medical supplies [1], locate particular patients [2], or guide visitors in long-term care homes [3]. In domestic environments, mobile robots find and retrieve household items [4], laundry [5], or cleaning [6]. In many such applications, reference images of the target object/person are provided a priori, enabling a robot to visually search for the object/person during navigation [7].

The aforementioned problem is known as Instance Image Goal Navigation (IIN). IIN is defined as a mobile robot requiring to navigate an unknown static environment to search for a person or object, given a reference image [8]. In IIN, the following challenges need to be addressed: 1) the viewpoint of the provided goal image may differ from the robot's viewpoint [9]; and 2) a robot must leverage visual and semantic context to improve navigation efficiency [10], [11].

To-date, existing IIN approaches have primarily used deep learning [7], [8], [12]-[23], and large foundation models [10], [24]-[26]. Deep learning methods have used convolutional encoders to extract features from visual observations and the goal image, capturing information such as scene geometry and visual appearance. These features are used to predict the robot's next action using actor-critic models [12]-[15], transformers [7], [16], diffusion models [17], recurrent neural networks [19] or inverse dynamics models [20]. Large foundation model approaches use pre-trained encoders from vision transformers to convert image observations and the goal image into semantic embeddings [10], [24]-[26]. The semantic embeddings are then used by scene graphs [10], [26], or reinforcement learning architectures [24], to direct the robot towards the goal. However, these existing methods assume that the goal image is captured from a viewpoint that is similar to the robot's viewpoint. For example, if the goal image is captured from a Bird's Eye View (BEV), the robot may fail to recognize it from its own frontal view perception due to feature differences across the two viewpoints [9]. For large foundation models, although high-level semantic embeddings from vision encoders are robust to viewpoint changes, category-level similarity is insufficient for IIN. IIN requires instance-level discrimination based on fine-grained visual and geometric cues that remain sensitive to viewpoint variation.

To address the aforementioned limitations, recent research has explored scene representations that can render novel viewpoints. In particular, 3D Gaussian Splatting (3DGS) [27] enables photorealistic 3D scene reconstruction using collections of 3D Gaussian ellipsoids. With respect to the IIN task, 3DGS, has been used for viewpoint rendering allowing a robot to localize and then navigate to the goal using waypoint-based planning [9], [28], [29], or model predictive control (MPC) [30]. However, these approaches assume that a detailed 3DGS map is available prior to robot deployment, which can limit their usability in unknown environments [31].

In this paper, we propose SplatSearch, a novel 3DGS-based architecture to address the IIN problem of a mobile robot searching for a static object or person in an unknown environment, using only a single reference image from an arbitrary viewpoint, Fig. 1. SplatSearch incrementally constructs a sparse 3DGS map from onboard RGB-D observations while detecting candidate objects. For each instance, novel viewpoints are rendered from the online map and completed using a view-consistent diffusion model to enable reliable feature matching against the goal image. Semantic context from RGB observations and visual context from synthesized views are fused to guide exploration. The robot navigates to the goal object if a matching instance is identified. The main contributions of SplatSearch are:

- The integration of a viewpoint synthesis method that renders multiple views for candidate objects using a sparse online 3DGS map.
- The development of View-Consistent Image Completion Network (VCICN), a multi-view diffusion model that inpaints missing regions in sparse 3DGS renders for feature matching with the goal image. This unique approach enables viewpoint-invariant goal recognition by combining geometry-based novel view rendering with diffusion-based inpainting
- A novel frontier exploration strategy. To the authors' knowledge, this strategy is the first to use: i) semantic context scores between RGB observations and the goal image, and ii) visual context scores from the inpainted novel viewpoints for frontier selection, allowing the robot to prioritize regions both semantically relevant and visually consistent with the target object/person.

## II. Related Works

We define existing IIN methods into: 1) deep learning-based methods [7], [8], [12]-[23], 2) large foundation model methods [10], [24]-[26], and 3) 3DGS methods [9], [28]-[30], [32].

### *A. Deep Learning-Based Methods*

Deep learning-based methods for the IIN task have mainly used deep reinforcement learning (DRL) [7], [8], [12]-[16], [19], [21]-[23] or imitation learning (IL) [17], [18], [20].

DRL-based methods have used convolutional encoders such as ResNet [7], [8], [12]-[15], [21], [22], graph convolutional networks [16], or vision transformers [19] to generate joint embeddings from RGB image observations and the goal image. DRL architectures including actor-critic networks [13]-[15], [22], proximal policy optimization (PPO) [8], [12], [21], [23], and transformers [7], [16] to generate navigation actions given the current and history of image observations. To enable multi-goal IIN tasks, navigation policies were memory-augmented using attention networks [12], [16], topological graphs [22], or feature-extraction encoders [13].

IL-based methods have generated robot navigation actions by learning from expert demonstrations by other robots using indoor navigation [17], [20], or IIN datasets [18]. In [17], IL was used to train a diffusion model to generate collision-free navigation trajectories, while being conditioned on the goal image. In [18], robots learned from demonstrations to maintain a topological graph of an environment, allowing them to generate subgoals towards the goal image. In [20], generative IL was used by training a variational encoder to predict the next observation, to reach the goal image.

### *B. Large Foundation Model Methods*

Large foundation models have either used vision-language models (VLMs) to generate semantic priors to locate a goal object [10], [24], or have queried a large language model (LLM) directly to inform the navigation policy [25], [26].

With respect to using VLMs, in [10], [24], RGB observations were converted into semantic embeddings using contrastive language image pretraining (CLIP) [33], while the goal image was processed through a ResNet encoder to locate the goal. For example, in [10], a semantic map of indoor environments was incrementally constructed using CLIP. In [24], the embeddings from the observations and goal were concatenated and trained with an actor-critic policy network.

Methods using LLMs to generate navigation actions have prompted the LLMs with an online scene graph of the environments, describing spatial and semantic relationships between objects [26], [25]. The scene graphs were used to generate plans towards candidate objects, and a local feature matching method, such as LightGlue [26] or SBERT [25], was used to identify the goal.

### *C. 3DGS Methods*

3DGS methods have leveraged novel view-synthesis to preserve detailed visual information across diverse viewpoints of a goal image [9], [28]-[30], [32]. Prior to navigation, these methods built high-fidelity 3DGS maps and embedded semantic features into the maps using CLIP [9], [29] or created topological graphs [28] for querying of goal images. In [9], multiple viewpoints around candidate objects were generated, then a feature matcher was used to identify the instance of the object. In [29], an improvement was made to [9], by estimating optimal viewpoints using hierarchical scoring. After an object instance is identified, navigation policies used to reach the goal object include diffusion [28], MPC [30] or the Fast Marching Method (FMM) [9], [29]. IGL-Nav, [32], builds an online 3DGS map without using a pre-built 3DGS representation. The 3DGS map is used for navigation by localizing the goal image pose.

### *D. Summary of Limitations*

DRL-based methods used pre-trained encoders to extract and compare features between a current observation and a goal image [7], [8], [12]-[16], [19], [21]-[23]. However, these encoders are not able to recognize goal objects from arbitrary viewpoints, resulting in degraded performance [9]. IL methods, [17], [18], [20], rely on trained policies from datasets, thus, their recognition ability is limited to the set of objects included in these datasets and they cannot generalize to new IIN scenarios [20]. The large foundation model methods, [10], [24]-[26], are unable to distinguish between visually distinct but semantically similar objects (e.g., two different chairs) [34]. 3DGS methods, [9], [28]-[30], assume that a dense 3DGS map is available to the robot prior to deployment, thus, cannot be applied to unknown environments. Closest to our work is IGL-Nav [32], which also builds on an online 3DGS map for visual navigation. However, the key differences are that: 1) IGL-Nav addresses image-goal navigation by localizing the goal camera pose, while SplatSearch addresses the IIN task by searching for a target object/person depicted in the goal image; and 2) IGL-Nav performs coarse-to-fine camera pose optimization using 3DGS map assuming sufficient reconstruction density, whereas SplatSearch renders object-centric novel viewpoints and applies diffusion-based inpainting to explicitly complete sparse 3DGS renderings for instance-level feature matching.

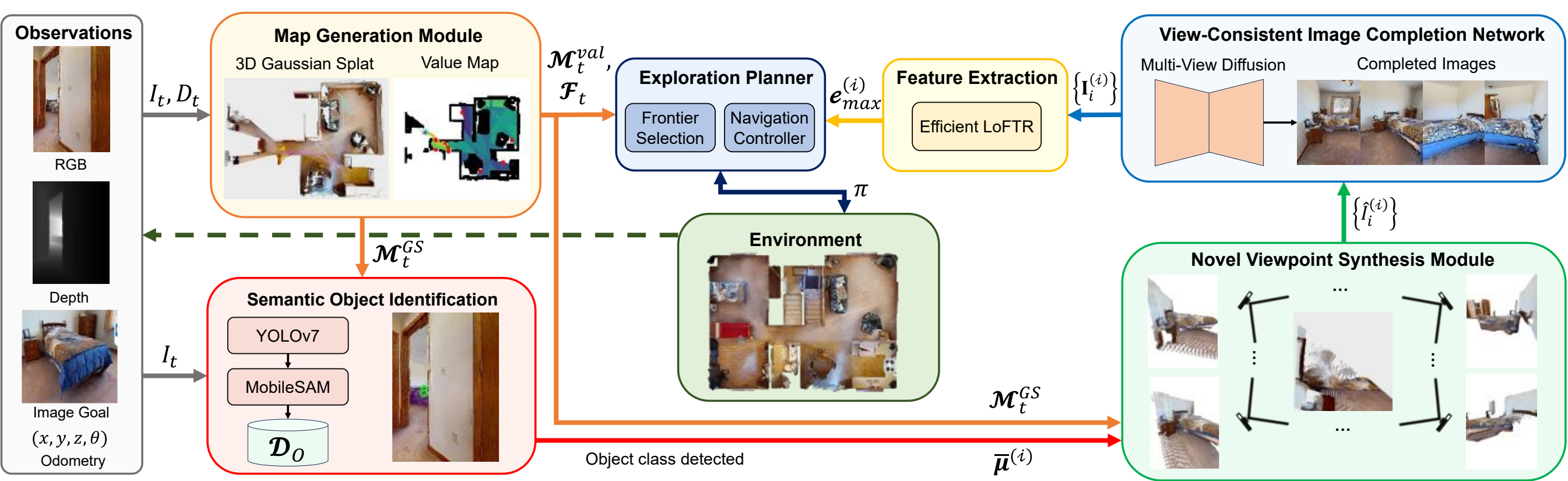

**Fig. 2.** The SplatSearch architecture, consisting of: 1) Map Generation Module (MGM) which generates a 3DGS map for photorealistic mapping and occupancy map for frontier exploration, 2) Semantic Object Identification Module (SOIM) detects if a current RGB image contains an object with the same class as the goal image, 3) Novel Viewpoint Synthesis Module (NVSM) generates multiple viewpoints around the object, 4) View-Consistent Image Completion Network (VCICN) inpaints the sparse viewpoint images from NVSM, 5) Feature Extraction Module (FEM) which uses the inpainted images to match features with the goal image, and 6) Exploration Planner (EP) which selects and generates feasible paths towards new frontiers.

To address the above limitations, we propose SplatSearch, the *first* online 3DGS-based architecture for IIN with goal images provided from arbitrary viewpoints in unknown environments. By constructing sparse 3DGS maps and inpainting viewpoints using a multi-view diffusion model, SplatSearch eliminates dependence on pre-built reconstructions, enabling deployment in unseen settings.

## III. Instance Image Goal Navigation in Unknown Environments

Robot IIN addresses the problem of a mobile robot, at an initial random pose in an unknown static 3D environment that needs to navigate to a goal object or person represented by an RGB image, $I^g \in \mathbb{R}^{H\times W\times 3}$, located at an unknown goal position $x^g$. At each timestep, $t$, the robot receives an RGB image and depth observation, $O_t = (I_t, D_t)$, from its onboard RGB-D camera, and executes an action, $a_t = \pi(O_t)$, according to a navigation policy, $\pi$. The total path length of the robot is: $d(\pi) = \sum_{t=0}^{T-1} \|x_{t+1} - x_t\|_2$, where $x_t$ is the robot position, and $T$ is the termination timestep. Robot navigation is considered successful if the robot reaches a distance within radius, $\tau_{succ}$, of the true goal position. The objective is to execute a navigation policy that minimizes the expected path length of the mobile robot subject to:

$$\pi^* = \underset{\pi}{\operatorname{argmin}}\, \mathbb{E}[d(\pi) \mid \|x_T - x^g\|_2 \le \tau_{succ}]. \qquad (1)$$

In IIN, the goal image serves as a visual reference for identifying the target object/person. The viewpoint of the goal image may be arbitrarily captured and is often physically unreachable by the robot due to embodiment and environment constraints (e.g., elevated or bird's-eye viewpoints) [8].

## IV. SplatSearch Architecture

The proposed SplatSearch architecture consists of six main modules, Fig. 2: 1) Map Generation Module (MGM), 2) Semantic Object Identification Module (SOIM), 3) Novel Viewpoint Synthesis Module (NVSM), 4) View-Consistent Image Completion Network (VCICN), 5) Feature Extraction Module (FEM), and 6) Exploration Planner (EP). Inputs are RGB-D observations with corresponding poses from the robot's onboard camera and a single goal image. These are first processed by the MGM, which incrementally constructs a sparse-view 3DGS map and a value map. The RGB-D observations and the 3DGS map are used as inputs into the SOIM to detect candidate objects of the same class as the goal image. The centroids of the detected objects are used by the NVSM to render multiple novel views using the 3DGS map. Since these renders are often incomplete due to sparse reconstructions, they are refined by the VCICN by inpainting missing regions to produce complete views. The completed renders are compared against the goal image by the FEM, outputting visual context scores that represent the likelihood of a detected instance matching the goal object. The EP uses the visual context scores and semantic scores obtained from a pre-trained image encoder to select frontiers for navigation. If a goal object is recognized, the EP plans a direct navigation path to its location. Each module is discussed in detail below.

Table I provides a summary of the key mathematical symbols and variables used in the SplatSearch architecture and corresponding equations.

### A. Map Generation Module (MGM)

The MGM utilizes $I_t$, $D_t$, and $x_t$ to construct: 1) a sparse-view 3DGS map of the environment for novel viewpoint rendering, and 2) a value map to generate semantic context scores to evaluate frontiers during exploration.

TABLE I: NOMENCLATURE OF KEY VARIABLES USED IN SPLATSEARCH

|  |  |
| --- | --- |
| $I_t, D_t$ | RGB and depth observations at timestep $t$ |
| $I^g$ | Goal RGB image |
| $x_t$ | Robot position at timestep $t$ |
| $x^g$ | True goal position |
| $\tau_{succ}$ | Success distance threshold |
| $G_i$ | $i$-th 3D Gaussian |
| $c_i, \mu_i, \Sigma_i, o_i$ | Color, mean position, covariance, opacity of Gaussian $G_i$ |
| $u_k$ | Pixel coordinate |
| $w_i(u_k)$ | Gaussian weight at pixel $u_k$ |
| $\hat{I}_t, \widehat{D}_t, \hat{O}_t$ | Rendered RGB, depth, and opacity images |
| $M_t^{GS}$ | Online sparse 3DGS map |
| $v_t$ | Semantic context score |
| $\bar{\boldsymbol{\mu}}^{(i)}$ | Estimated 3D centroid of object instance $i$ |
| $N_R$ | Number of novel viewpoints |
| $r_v$ | Sampling radius |
| $z_t^i$ | Latent diffusion representation of viewpoint $i$ |
| $s_{\max}^{(i)}$ | Maximum feature matching score for object $i$ |
| $F_t$ | Frontier set |
| $\bar{V}_t(f)$ | Average semantic value at frontier $f$ |

*1) 3DGS Map*

Following ActiveSplat [35], we maintain an incremental sparse-view 3DGS map for scene representation. A 3DGS map of the environment at timestep $t$, $\boldsymbol{\mathcal{M}}_t^{GS}$, is represented by anisotropic 3D Gaussians, $G_i$, each parameterized by color $\boldsymbol{c}_i$, position $\boldsymbol{\mu}_i \in \mathbb{R}^3$, covariance $\boldsymbol{\Sigma}_i \in \mathbb{R}^{3\times3}$, and opacity $\boldsymbol{o}_i$. $\mathcal{P} : \mathbb{R}^2 \rightarrow \mathbb{R}^3$ denotes the mapping from a 2D pixel coordinate in the RGB image, $\boldsymbol{u}_k \in \mathbb{R}^2$, to its corresponding 3D point in the scene. The weight of $G_i$, at pixel $\boldsymbol{u}_k$ in the rendered image is given by [35]:

$$w_i(\boldsymbol{u}_k) = \boldsymbol{o}_i \exp\left(-\frac{1}{2}(\mathcal{P}(\boldsymbol{u}_k) - \boldsymbol{\mu}_i)^T \boldsymbol{\Sigma}_i^{-1}(\mathcal{P}(\boldsymbol{u}_k) - \boldsymbol{\mu}_i)\right), \quad (2)$$

where $i \in \{1, \ldots, N_G\}$ represents indices for the set of 3D Gaussians and $k \in \{1, \ldots, N_P\}$ represents the indices for the set of pixels in the RGB image. The rendered RGB color $\hat{I}_t$, depth $\widehat{D}_t$ and opacity $\hat{O}_t$, are generated by alpha-compositing the 2D projection of each Gaussian into the image plane [36]. The 3DGS map is optimized online by minimizing both the photometric loss, $\mathcal{L}_p$, and geometric loss, $\mathcal{L}_g$ [35]:

$$\mathcal{L}_p = |I_t - \hat{I}_t| + \left(1 - \mathrm{SSIM}(I_t, \hat{I}_t))\right), \quad (3)$$

$$\mathcal{L}_g = |D_t - \widehat{D}_k|. \quad (4)$$

The Structural Similarity Index (SSIM) is used to minimize perceptual differences between the rendered and observed RGB images. The 3DGS map is updated online at each timestep in regions which are not previously mapped or with low accumulated opacity:

$$M_t(u_k) = (\hat{O}_t < \tau_o) \cup ((D_t < \widehat{D}_t) \cap (|D_t - \widehat{D}_t| > \tau_M)), \quad (5)$$

where $\tau_o$ is the opacity threshold, and $\tau_M$ is the median depth error. Pixels satisfying $M_t(u_k) = 1$ are used to update the corresponding Gaussians in the 3DGS map.

*2) Value Map*

The depth images, $D_t$, are used to build an online occupancy map, $\boldsymbol{\mathcal{M}}_t^{occ}$. Frontier points, $\boldsymbol{\mathcal{F}}_t$, are extracted as the boundary locations between the free and unknown cells. The value map, $\boldsymbol{\mathcal{M}}_t^{val}$, is constructed by assigning each free-space cell in $\boldsymbol{\mathcal{M}}_t^{occ}$, with a semantic context score quantifying its visual and semantic relevance to $I^g$. To obtain the semantic context scores, we use *only the image encoder* of a pre-trained BLIP-2 VLM [37] to generate semantic embeddings for the goal image, $e_G$, and the current RGB observation, $e_t$. The semantic score is determined through the embedding cosine similarity: $v_t = \frac{e_t \cdot e_g}{\|e_t\| \|e_g\|}$, where a higher value represents a higher semantic relevance to the goal image. The value $v_t$ is stored in all of the free space pixels within the current field-of-view (FOV) of the robot.

We also generate a confidence map, $\boldsymbol{\mathcal{M}}_t^{conf}$, to update a pixel's semantic value based on its assigned value and the current FOV of the robot following [11]. For each frontier, $f \in \boldsymbol{\mathcal{F}}_t$, we compute a normalized average semantic score, $\bar{V}_t(f)$ as the mean of $\boldsymbol{\mathcal{M}}_t^{val}$ within a circle of radius $r_{val}$ around the frontier point, $f$. These scores are used by the EP to guide frontier selection towards regions that are semantically similar to the goal image.

## B. *Semantic Object Identification Module (SOIM)*

The SOIM determines whether the current RGB image, $I_t$, contains an instance belonging to the image goal class, and assigns a unique identifier to each detected object. We use YOLOv7 [38] to generate bounding boxes around candidate objects and MobileSAM [39] to generate instance masks $\{m_{t,n}\}_{n=0}^{n_t}$, where $n_t$ denotes the number of masks detected at timestep $t$. For each mask $m_{t,n}$, the pixels are mapped to their corresponding 3D Gaussians in the 3DGS map, $\boldsymbol{\mathcal{M}}_t^{GS}$, using $\mathcal{P}$. The corresponding set of Gaussians is denoted as $\boldsymbol{\mathcal{G}}^{(i)} = \{\boldsymbol{\mu}_i^{(i)}\}$, which are aggregated and averaged to estimate the 3D centroid of the object, $\bar{\boldsymbol{\mu}}^{(i)}$, with identification $i$. Each object is stored in an object database $\boldsymbol{\mathcal{D}}_O$ as: $\boldsymbol{\mathcal{O}}^{(i)} = \{\boldsymbol{\mathcal{G}}^{(i)}, \bar{\boldsymbol{\mu}}^{(i)}, \boldsymbol{\mathcal{X}}^{(i)}\}$, where $\boldsymbol{\mathcal{X}}^{(i)}$ represents the set of positions from which object $i$ has been observed. When a new object is detected, its centroid $\bar{\boldsymbol{\mu}}^{new}$ is compared to existing object centroids in $\boldsymbol{\mathcal{D}}_O$. If no match is found, it is appended as a new entry to $\boldsymbol{\mathcal{D}}_O$; otherwise, the matched object is updated with the new viewpoint. The centroid, $\bar{\boldsymbol{\mu}}^{(i)}$, is passed to the NVSM each time an object in $\boldsymbol{\mathcal{D}}_O$ has been observed in $f_O$ new distinct viewpoints (i.e. when $|\boldsymbol{\mathcal{X}}^{(i)}| \bmod f_O = 0$ ). This ensures the NVSM is only executed when increasingly refined object information is obtained.

## C. *Novel Viewpoint Synthesis Module (NVSM)*

The NVSM generates camera viewpoints using the 3DGS map and the input centroid, $\bar{\boldsymbol{\mu}}^{(i)}$, to render novel images around the object. Around the centroid, $\bar{\boldsymbol{\mu}}^{(i)}$, viewpoints are sampled along aspherical surface of radius $r_v$, restricted to the upper hemisphere to avoid occlusions by the ground plane. During this sampling process, we used the FEM (Section IV.E) to compare each newly generated viewpoint with the previously generated viewpoint. If there were no feature correspondences between the previous viewpoint and the current viewpoint, an occlusion is likely to have happened and the current viewpoint is discarded. A total of $N_R = 8$ viewpoints are generated by rotating at fixed angular steps, with each camera-to-world transformation matrix, $\mathbf{T}_{c\to w}$, positioned tangentially to the sphere and oriented towards the centroid. Using these transformation matrices, the set of novel images, $\left\{\hat{I}_i^{(i)}\right\}_{i=0}^{N_R}$, are rendered using the 3DGS map, and are subsequently provided to the VCICN.

D. *View-Consistent Image Completion Network (VCICN)*

The VCICN is used to complete rendered images by inpainting pixels that are unobserved or partially observed in the 3DGS map. It uses a multi-view diffusion network to generate structurally consistent and complete images that can be reliably compared by the FEM, since the 3DGS map is only sparsely generated. The input to the VCICN consists of $N_R$ viewpoints, $\left\{\hat{I}_i^{(i)}\right\}_{i=0}^{N_R}$, and their corresponding binary masks $\left\{\widehat{M}_i^{(i)}\right\}_{i=0}^{N_R}$, representing the observed and missing regions of the image. We set $\hat{I}_0^{(i)}$ as the robot view in which the candidate object is detected, and $\widehat{M}_0^{(i)} = \mathbf{0}$, so that the first viewpoint serves as the unmasked reference image that diffusion inpainting is conditioned on [40].
Each image is encoded into a latent feature map using a variational autoencoder (VAE): $z_0^i = \mathrm{VAE}(\hat{I}_i^{(i)}) \in \mathbb{R}^{H\times W\times 4}$. The binary mask $\widehat{M}_i^{(i)}$ is down-sampled using the same VAE network to match the latent resolution. For each latent $z_0^i$, we simulate the forward diffusion process at a randomly chosen timestep $t$ to produce a noisy latent vector:

$$z_t^i = \sqrt{\alpha_t} z_0^i + \sqrt{1-\alpha_t}\epsilon, \qquad (6)$$

where $\alpha_t$ follows a consistent noise schedule, defining the amount of noise added at each timestep, and $\epsilon \sim \mathcal{N}(0, \mathbf{I})$. The sequence of inputs at timestep $t$ is:

$$\boldsymbol{x}_t = \left[z_t^{0:N_R}, z^{0:N_R}\left(1-\widehat{M}^{0:N_R}\right), \widehat{M}^{0:N_R}\right], \qquad (7)$$

where $\widehat{M}$ is the down sampled mask. The VCICN predicts the noise $\epsilon_\theta(x_t, \tau_\phi(c), t)$, where $c$ is the CLIP-encoded text prompt corresponding to the goal object. The network is trained with an MSE loss [40]:

$$\mathcal{L}_D = \mathbb{E}_{x_0,\epsilon,t,c}\left[\left\|\epsilon - \epsilon_\theta\left(\boldsymbol{x}_t, \tau_\phi(c), t\right)\right\|_2^2\right]. \qquad (8)$$

After reverse diffusion denoising, the VCICN outputs the completed viewpoint set of images, $\left\{\mathbf{I}_i^{(i)}\right\}_{i=0}^{N_R}$, which are then used by the FEM for feature extraction against the goal image.

E. *Feature Extraction Module (FEM)*

The FEM determines whether the current object instance corresponds to the goal image, $I^g$. The input is the set of completed novel viewpoints $\left\{\mathbf{I}_i^{(i)}\right\}_{i=0}^{N_R}$, generated from the VCICN. The object score is computed using Efficient LoFTR [41], denoted by $\boldsymbol{E}(\cdot,\cdot)$, which returns the number of feature correspondences between two images. Efficient LoFTR is selected due to its efficiency and accuracy for image matching when viewpoint changes and partial visual artifacts are present. For each viewpoint, $\mathbf{I}_i^{(i)}$, the viewpoint with the highest match is computed as: $i^* = \underset{i}{\operatorname{argmax}}\, \boldsymbol{E}\left(I^g, \mathbf{I}_i^{(i)}\right)$. If the object score satisfies $\boldsymbol{s}_{max}^{(i)} = \boldsymbol{E}\left(I^g, \mathbf{I}_{i^*}^{(i)}\right) \geq \tau_m$, where $\tau_m$ is the feature-match threshold determined through simulated experiments, the object is declared to match the goal instance and is passed into the EP for selection of the final frontier. If $\boldsymbol{s}_{max}^{(i)} < \tau_m$, the object is not considered a match and the score is used by the EP to determine the next frontier.

F. *Exploration Planner (EP)*

The EP utilizes a frontier exploration strategy. Our contribution to traditional frontier exploration is the combined use of: 1) semantic context from the value map to enable the EP to prioritize frontiers that are semantically relevant to the goal image, and 2) visual context from the FEM to select frontiers close to object instances with higher feature-matching scores. Namely, the EP determines the next frontier location to navigate to using $\overline{V}_t(f)$ and $\boldsymbol{\mathcal{F}}_t$ from the MGM, and $\left\{\boldsymbol{s}_{max}^{(i)}\right\}_{i=0}^{|\boldsymbol{\mathcal{D}}_O|}$ from the FEM. During exploration, if the FEM has not detected the goal object, the EP selects from the set of frontiers, $f \in \boldsymbol{\mathcal{F}}_t$, that maximizes the utility function:

$$f_t^* = \underset{f\in\boldsymbol{\mathcal{F}}_t}{\operatorname{argmax}}[\alpha \cdot d_t(f) + \beta \cdot \mathcal{E}_t(f) + \delta \cdot \overline{V}_t(f)\,], \qquad (9)$$

where $\alpha$, $\beta$, $\delta$ are the weighting coefficients, $d_t(f)$ is the normalized inverse distance between $x_t$ and $f$. $\mathcal{E}_t(f)$ is the normalized object score determined as:

$$\mathcal{E}_t(f) = \frac{1}{C}\max_i \frac{\boldsymbol{s}_{max}^{(i)}}{\tau_m \|f - \overline{\boldsymbol{\mu}}^{(i)}\|}, \qquad (10)$$

where $C$ is the normalization constant. $\boldsymbol{\mathcal{M}}_t^{occ}$ is then used by the Navigation Controller, which utilizes (FMM) [42] to generate the shortest feasible path to navigate from the current location, $x_t$, to $x_T = f_t^*$. If the FEM confirms a match with the goal image, FMM plans a path to the estimated 3D centroid $\overline{\boldsymbol{\mu}}^{(i)*}$ of the matched instance. Navigation terminates when the robot reaches a position within a predefined spatial tolerance $\tau_{succ} = 1$m of this centroid, which is consistent with commonly adopted success thresholds in instance-level and object-centric embodied navigation benchmarks (e.g., [8]), at which point the EP issues a stop action and the episode is completed.

## V. Multi-View Diffusion Inpainting

The details of data collection and training for the VCICN are explained below.

A. *Dataset Collection*

We used the Habitat simulator with the HM3DSem-v0.2 dataset [43]. To construct the training data for VCICN, we executed SplatSearch (without VCICN) on the 145 environments in the training set, spanning six object classes (TV, sofa, chair, table, plant, monitor) [8]. When the SOIM detects an object instance, we applied the NVSM to capture $N_R = 8$ viewpoints around the object, with their corresponding masks representing regions that are not constructed. This process produced a dataset of approximately 2,000 objects, corresponding to 16,000 images in total. We augment this dataset with 400 people each with 8 viewpoints from the real-world EgoBody dataset [44], totaling 19,200 images.

TABLE II: COMPARISON BETWEEN SPLATSEARCH AND IIN SOTAS

| Method | HM3D-val | | HM3D-val-hard | |
|---|---|---|---|---|
| | SR ↑ | SPL ↑ | SR ↑ | SPL ↑ |
| Mod-IIN | 0.5610 | 0.2330 | 0.2200 | 0.1354 |
| GaussNav | 0.5400 | 0.0673 | 0.4900 | 0.0428 |
| IEVE | 0.6600 | 0.2735 | 0.5800 | 0.1932 |
| UniGoal | 0.6900 | 0.2837 | 0.5900 | 0.1738 |
| **SplatSearch** | **0.7000** | **0.3740** | **0.6300** | **0.2675** |

## B. *Training*

Training was completed on an RTX 4090 GPU. The model was initialized using the weights of MVInpainter-O [40] (which is trained on large-scale real-world multi-view datasets), it was fine-tuned for 10k steps with a batch size of 64, learning rate of 0.0001 on a curated dataset consisting of simulated and real-world samples. Training was completed in 46 hours using early stopping.

# VI. EXPERIMENTS

We evaluated SplatSearch by conducting: 1) a comparison study with state-of-the-art (SOTA) methods, 2) an ablation study to investigate the design choices, and 3) real-world experiments to evaluate the generalizability of SplatSearch in real-world environments.

## A. *Comparison Study*

We evaluated SplatSearch in the Habitat simulator across 100 episodes using two validation sets, Fig. A.1 (in the supplementary materials): 1) the standard HM3DSem-v0.2 validation dataset (HM3D-val) [8], and 2) a modified dataset designed to evaluate robustness to viewpoint invariance, HM3D-val-hard. We construct HM3D-val-hard by re-sampling goal images from HM3D-val in spherical coordinates around the object centroid with radial distance $r \in [0.5, 2.5]m$, azimuth $\theta \in (0, \pi]$, elevations, $z \in [1.0, 2.0]m$, and camera angle further perturbed between $[-10^o, +10^o]$. The performance metrics used to evaluate each benchmark are: 1) the Success Rate (SR); measures the percentage of episodes where the robot successfully reaches the goal object, and 2) the success path length (SPL); measures the robot navigation efficiency: $SPL = \frac{1}{N_T}\sum_{i=1}^{N_T} S_i \frac{d(\pi^*)}{\max(d(\pi), d(\pi^*))}$ where $N_T$ is the number of trials.

### 1) *Comparison Methods:*

We compared SplatSearch with the SOTA methods:

**Mod-IIN** [45]: A modular classical method that uses frontier exploration, keypoint-based feature extraction and matching for object reidentification, and FFM for local navigation. This was selected as a representative classical modular method.

**GaussNav** [9]**:** Uses a pre-built 3DGS map to generate viewpoints of goal-class instances, a local feature matcher for goal localization, and a classical planner for navigation. It was selected as a representative 3DGS approach.

**IEVE** [23]**:** A DRL approach that uses a switch policy to decide either explore a new frontier or verify current observation against the goal image using classical feature matching. IEVE was selected as it is the SOTA learning-based approach for IIN.

**UniGoal** [26]**:** Uses an LLM to build an online scene graph in semantic space, and a classical feature matcher to compare the goal image against the RGB observation. UniGoal uses *only semantic context* provided by the scene graph and LLM to evaluate frontiers for navigation.

2) *Results:* The SR and SPL of SplatSearch and the SOTA comparison methods are presented in Table II. SplatSearch achieved the highest SR and SPL with respect to the SOTA methods for both datasets. In general, the overall performance of all the methods degraded for HM3D-val-hard dataset due to the increased challenge of recognizing the goal image from BEV perspectives. However, SplatSearch was able to render objects from any viewpoint using the NVSM and VCICN, resulting in robust feature matching correspondences for recognition of the specific instance of the goal image. Mod-IIN achieved lower SR and SPL than SplatSearch, particularly on HM3D-val-hard, as it relies on greedy frontier exploration based on geometric coverage, and does not leverage semantic and visual context. As a result, Mod-IIN often explored large portions of the environment that were irrelevant to the target object before encountering the goal. As IEVE does not generate a 3D scene representation and is unable to match features from the goal image to new unseen viewpoints, it had lower SRs and SPLs. UniGoal had lower SPL and SR than SplatSearch, as it converts the goal image into semantic labels and must physically navigate to candidate objects to perform RGB-based feature matching. GaussNav achieved the lowest SRs and SPLs as it requires constructing the 3DGS map of the full environment prior to search.

## B. *Ablation Study*

We conducted an ablation study with different variants of SplatSearch on the HM3D-val and HM3D-val-hard datasets.

**SplatSearch without (w/o) NVSM:** Uses $I_t$ to compute visual context scores instead of rendering novel viewpoints. It evaluates the impact of viewpoint synthesis.

**SplatSearch w/o VCICN:** Uses the NVSM to generate novel viewpoints, however, does not use multi-view diffusion. This determines the effect of inpainting on feature matching.

**SplatSearch w/o Semantic Context Scores (SCS):** Does not include SCS, namely $\overline{V_t}(f)$. This determines the effect of semantic context scores on frontier selection.

**SplatSearch w/o Visual Context Scores (VCS):** Does not include VCS, namely $\mathcal{E}_t(f)$ in Equation 14. Determines the effect of visual context scores on frontier selection.

**SplatSearch ($N_R$ = 4, 12):** Evaluate the impact of varying the number of rendered viewpoints.

**SplatSearch ($r_v$ = 0.5, 2):** Evaluate the effect of varying the sampling radius.

***1)*** *Results:* The results for the ablation study are presented in Table III. Overall, SplatSearch achieved the highest SR and SPL across all of the variants. SplatSearch w/o VCICN had a lower SR and SPL, as the rendered images contained large missing regions, which degraded the performance of FEM. SplatSearch w/o NVSM also had a lower SR and SPL as it does not render candidate objects from arbitrary viewpoints.

SplatSearch w/o SCS achieved a lower SPL, which highlights the importance of the value map in identifying semantic relationships between objects to select frontiers. SplatSearch w/o VCS did not select frontiers that were nearby objects with high feature matches, resulting in redundant exploration. SplatSearch ($N_R = 4$) achieved lower SR and SPL as the decreased angular coverage around the object lowered the probability of viewpoint alignment with the goal image, reducing consistent correspondences in the FEM. SplatSearch ($N_R = 12$) did not improve performance. Although viewpoint coverage increased, the larger multi-view input increased memory consumption of the VCICN. This resulted in the FEM operating on incomplete renderings without diffusion-based inpainting. SplatSearch ($r_v = 0.5$) produced zoomed-in renderings with reduced contextual structure and fewer stable features, whereas SplatSearch ( $r_v = 2.0$ ) increased the likelihood of partial occlusions from intervening scene elements. The latter variant reduced viewpoint similarity to the goal image, thereby lowering correspondence quality. These results highlight the importance of the NVSM and VCICN for identifying goal images from arbitrary viewpoints, and the SCS and VCS for frontier selection. We emphasize the effect of VCICN by presenting a qualitative visualization of the sparse novel views and their diffusion-inpainted results in Fig. B.1 (in the supplementary document).

### C. *Real-World Experiments*

We conducted real-world experiments in a 3-room indoor space (size 35m x 14m) at a university campus, Fig. 3. A Jackal robot with an onboard ZED2 stereo camera was used. The robot pose was estimated using LF-SLAM [46], which jointly optimizes camera poses and 3D line landmarks by minimizing the reprojection error between observed 2D line segments and their corresponding 3D map lines. No additional training was used for the new real-world deployment. SplatSearch was run on a workstation with an NVIDIA RTX 3090 GPU, an AMD Ryzen 7 5800X CPU and 64 GB RAM, and communicated with Jackal via Wi-Fi connection. We evaluated five image goals: a person, TV, chair, backpack, and laptop, which included objects in the simulation dataset (TV, chair) and new objects not in the simulation dataset (backpack, laptop). We randomized the reference image viewpoints across frontal, side, and BEV perspectives, for a total of 25 trials. We compared SplatSearch with SplatSearch w/o VCIVN variant to investigate the impact of VCICN in real-world scenarios. SplatSearch achieved an SR = 0.76 and SPL = 0.48 across all trials, while SplatSearch w/o VCICN achieved an SR = 0.60 and SPL = 0.36. The degraded performance of SplatSearch w/o VCICN shows that sparse real-world 3DGS reconstructions can produce incomplete views which degrade feature correspondences. VCICN performs explicit completion of missing regions in the rendered views for object feature matching under sparse real-world constructions. SplatSearch achieved comparable performance as the simulation results. VCICN was able to generalize to real-world scenarios, as: 1) the pretrained MVInpainter was trained on large-scale real-world multi-view datasets [40], and 2) VCICN was finetuned on the real-world human motion dataset EgoBody [44].

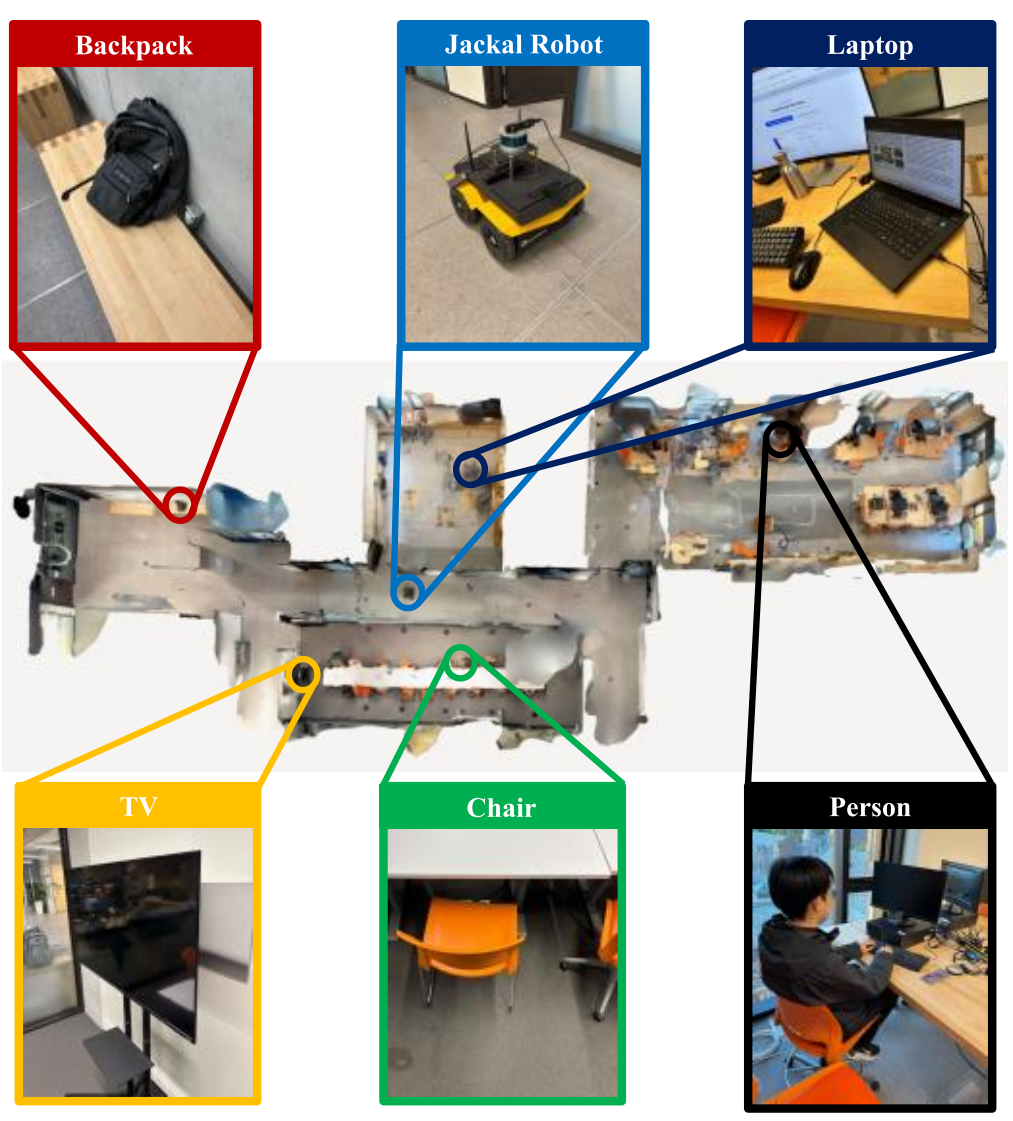

**Fig. 3.** SplatSearch was deployed in a real-world indoor environment with goal images of five new objects including a person, chair, laptop, TV and backpack.

TABLE III: ABLATION STUDY

| Variants | HM3D-val | | HM3D-val-hard | |
|---|---|---|---|---|
| | **SR ↑** | **SPL ↑** | **SR ↑** | **SPL ↑** |
| SplatSearch w/o NVSM | 0.6200 | 0.2863 | 0.4300 | 0.1712 |
| SplatSearch w/o VCICN | 0.6500 | 0.3174 | 0.5800 | 0.2204 |
| SplatSearch w/o SCS | 0.6400 | 0.3155 | 0.5600 | 0.2489 |
| SplatSearch w/o VCS | 0.6700 | 0.3407 | 0.5900 | 0.2476 |
| SplatSearch ($N_R = 4$) | 0.5500 | 0.2338 | 0.4700 | 0.1847 |
| SplatSearch ($N_R = 12$) | 0.6700 | 0.3684 | 0.6000 | 0.2512 |
| SplatSearch ($r_v = 0.5$) | 0.5900 | 0.2198 | 0.5100 | 0.2086 |
| SplatSearch ($r_v = 2.0$) | 0.6100 | 0.2280 | 0.5500 | 0.2148 |
| **SplatSearch** | **0.7000** | **0.3740** | **0.6300** | **0.2675** |

We additionally conducted a runtime analysis on the modules in SplatSearch. The results are shown in Table C.I in the supplementary materials. Although the multi-step diffusion of VCICN takes several seconds (~7s), it is invoked only when a candidate object instance has been detected and observed by the robot from $f_O$ new distinct viewpoints (Section IV.B), it does not run continuously. VCICN also runs in parallel with the main navigation loop (MGM, SOIM and EP), whose combined runtime is 0.21s, thus enabling practical real-time deployment. A video of our results and supplementary materials document are shown on our project webpage, https://anon-splatsearch.github.io/.

As SplatSearch uses YOLOv7 [38], it is limited to a fixed set of detectable object categories, this can be addressed by replacing YOLOv7 with an open-vocabulary detector such as Grounding DINO [47]. As a multi-module system similar to other IIN methods [9], [23], [45], failure of individual processing components (e.g., detection, mapping), can affect overall performance. This can be addressed by incorporating a backup strategy that does not rely on the failed module. For example, if the NVSM or VCICN fails, the system can bypass them and perform feature matching directly using the robot's current RGB observations.

## VII. CONCLUSION

In this paper, we presented SplatSearch, a novel architecture to address the IIN problem for mobile robots in unknown environments. SplatSearch uniquely uses a combination of sparse-view 3DGS reconstructions to synthesize candidate viewpoints around objects, and a multi-view diffusion model to inpaint missing regions of the images for goal recognition across arbitrary viewpoints of candidate objects or people. A frontier policy combines visual and semantic context from onboard RGB observations to guide the robot towards the goal image. Extensive photorealistic experiments validate SplatSearch's performance when compared to SOTA methods. An ablation study validated our design choices for the NVSM for novel viewpoint rendering and the VCICN for diffusion-based image inpainting in our architecture. Real-world experiments show the ability of SplatSearch to perform IIN for new and unseen objects in a new environment. Future work will extend our SplatSearch architecture to use open-vocabulary object detector [47] and consider dynamic obstacles in environments.